\documentclass[runningheads]{llncs}

\usepackage{booktabs}   
\usepackage{subcaption} 

\usepackage[utf8]{inputenc} 
\usepackage[T1]{fontenc}    
\usepackage{hyperref}
\usepackage{url}            
\usepackage{booktabs}       
\usepackage{amsfonts}       
\usepackage{nicefrac}       

\usepackage{graphicx} 
\usepackage{pgfplots} 
\usepackage{tikz} 
\usetikzlibrary{shapes,arrows}
\usepackage{pgfplots}
\usepackage{float}
\usepackage{bussproofs}
\usepackage{xspace}
\usepackage{amsmath}
\usepackage{amssymb}
\usepackage{mathtools}
\usepackage{doi}
\usepackage{alltt}

\def\holfour{\textsf{HOL4}\xspace}

\usepackage{color}

\usetikzlibrary{arrows,shapes,calc}
\usetikzlibrary{trees,positioning,fit}
\tikzstyle{arrow}=[draw,-to,thick]
\tikzstyle{embedding} = [draw, minimum width=15mm, minimum height=6mm]
\tikzstyle{nnop} = [draw, minimum width=8mm, minimum height=8mm, rounded 
corners]
\tikzstyle{nnopbig} = [draw, minimum width=15mm, minimum height=15mm, rounded 
corners]
\tikzstyle{block} =
  [rectangle, draw,
   minimum width=9em, text centered, rounded corners, minimum height=2.0em]
\tikzstyle{line}=[draw]
\tikzstyle{cloud} =
  [draw, text centered, ellipse, minimum height=2.0em, minimum width=9em]

\makeatletter
\renewcommand\section{\@startsection{section}{1}{\z@}%
                       {-12\p@ \@plus -4\p@ \@minus -4\p@}%
                       {8\p@ \@plus 4\p@ \@minus 4\p@}%
                       {\normalfont\large\bfseries\boldmath
                        \rightskip=\z@ \@plus 8em\pretolerance=10000 }}
\makeatother

\begin{document}

\title{Tree Neural Networks in HOL4\thanks{This work  has been supported 
by the European Research Council (ERC) grant 
AI4REASON no. 649043  under the EU-H2020 programme. 
We would like to thank 
Josef Urban for his contributions to the final version of this paper.}}
\author{Thibault Gauthier}
\institute{Czech Technical University in Prague, Prague, Czech Republic\\
\email{email@thibaultgauthier.fr}}
\authorrunning{T. Gauthier}
\titlerunning{Tree Neural Networks in HOL4}
\maketitle    
          
\begin{abstract}
We present an implementation of tree neural networks within the 
proof assistant HOL4. Their architecture makes them naturally suited for 
approximating functions whose domain is a set of formulas.
We measure the performance of our implementation and compare it with 
other machine learning predictors on the tasks of evaluating arithmetical 
expressions and estimating the truth of propositional formulas.
\end{abstract}

\section{Introduction}
Applying machine learning to improve proof automation has been an essential
topic in the theorem proving community and contributed to the rise of powerful 
automation such as hammers~\cite{hammers4qed}.
In these systems, the current machine learning predictors learn the premise 
selection task with relative success. However, these predictors typically rely 
on a set of syntactic features, and thus, they can hardly discover semantic 
patterns. To solve this 
issue, we propose in this work to rely on deep learning models to automatically 
infer appropriate features that better approximates object semantics.
The success of this approach depends heavily
on how the design of the neural network architecture encodes and 
processes the input objects. For example, the space invariance of convolutional 
neural networks makes them successful at interpreting images. Moreover, 
recurrent networks can process arbitrarily long sequences of tokens, 
which is necessary for learning text-based tasks.
In the case of formulas, tree neural 
networks(TNNs)~\cite{DBLP:journals/tacl/KiperwasserG16a}
capture the compositional nature of the underlying functions as 
their structure dynamically imitates the tree structure of the formula 
considered.

That is why we implement TNNs in \holfour~\cite{hol4} and 
evaluate their pattern recognition abilities on two tasks related to theorem 
proving.
The first task is to estimate the value of an 
expression. It is an example of evaluating a formula in a Tarski-style model,
which can be in general useful for conjecturing and approximate reasoning.
The second task is to estimate the truth of a formula. 
Acquiring this ability is important for discarding false conjectures and 
flawed derivations.
These two tasks are only a sample of the many theorem proving tasks that could 
be learned. We believe that deep learning 
models such as TNNs could be useful to guide automated theorem provers. In 
practice, the existence of an implementation of a deep learning predictor in 
\holfour is a valuable tool for improving proof automation methods in this 
proof assistant.
Experiments on the implemented TNNs presented in this paper can be replicated 
by following the instructions in the 
file~\footnote{\url{HOL/examples/AI_TNN/README.md}}
from the \holfour 
repository~\footnote{\url{https://github.com/HOL-Theorem-Prover/HOL}}
after switching to this
commit~\footnote{c679f0c69b397bede9fefef82197f33ec495dd8a}.

\section{Tree Neural Networks}\label{sec:tnn}
\begin{figure}[h]
\centering
\vspace{-5mm}
\begin{tikzpicture}[scale=0.8, every node/.style={scale=0.8}]
\node [embedding,node distance=3cm] (0l) {$0$};
\node [node distance=1.5cm,right of=0l] (0lm) {};
\node [embedding, node distance=3cm, right of=0l] (0m) {$0$};
\node [node distance=3cm, right of=0m] (0r) {};
\node [embedding, node distance=1.5cm, right of=0r] (0rr) {$0$};
\node [nnop, node distance=0.5cm, above of=0lm] (p) {$\times$};
\node [nnop, node distance=0.5cm, above of=0r] (s) {$s$};
\node [node distance=1cm, above of=s] (se) {};
\node [embedding, node distance=1cm, left of=se] (ser) {$s(0)$};
\node [embedding, node distance=1cm, above of=p] (pe) {$0 \times 0$};
\node [node distance=1.75cm, right of=pe] (pr) {};
\node [nnop, node distance=0.5cm, above of=pr] (t) {$+$};
\node [embedding, node distance=1cm,above of=t] (te) {$0 
\times 0 + s(0)$};
\node [nnop, node distance=1cm,above of=te] (h) {$\mathit{head}$};
\node [node distance=1cm,above of=h] (he) {};
\draw[-to,thick] (0l) to (p);
\draw[-to,thick] (0m) to (p);
\draw[-to,thick] (0rr) to (s);
\draw[-to,thick] (p) to (pe);
\draw[-to,thick] (s) to (ser);
\draw[-to,thick] (pe) to (t);
\draw[-to,thick] (ser) to (t);
\draw[-to,thick] (t) to (te);
\draw[-to,thick] (te) to (h);
\draw[-to,thick] (h) to (he);
\end{tikzpicture}
\caption{Computation flow of a tree neural network on the arithmetical 
expression $0 \times 0 + s(0)$. The operator $s$ stands for the successor 
function.  Rectangles represent embeddings (in $\mathbb{R}^d$) and 
rounded squares represent neural networks.}\label{fig:tnn}.
\vspace{-10mm}
\end{figure}
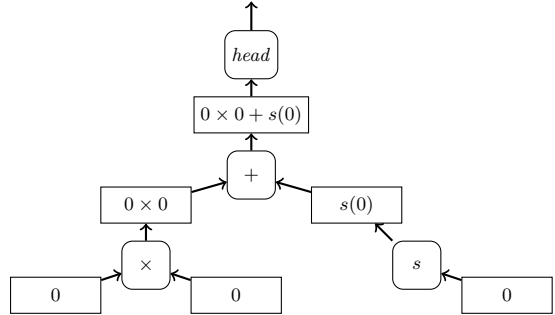

Let $\mathbb{O}$ be a set of operators (functions and constants) and 
$\mathbb{T}_\mathbb{O}$ be all terms built from operators in $\mathbb{O}$. 
A TNN is to approximate a function from $\mathbb{T}_\mathbb{O}$ to 
$\mathbb{R}^n$.
A TNN consists of a head network $N_\mathit{head}$ and a mapping that 
associates to each operator $f \in \mathbb{O}$ a neural network $N_f$.
If the operator $f$ has arity $a$, it is to learn a function from 
$\mathbb{R}^{a \times d}$ to $\mathbb{R}^d$. 
And the head network $N_\mathit{head}$ is to approximate a function from 
$\mathbb{R}^d$ to $\mathbb{R}^n$.
The natural number $d$ is called the \textit{dimension} of the embedding space.
For a TNN, an embedding function $E: \mathbb{T}_\mathbb{O} \mapsto 
\mathbb{R}^d$ can be recursively defined by: 
\[E(f(t_1,\ldots,t_a))=_\mathit{def} N_f(E(t_1),\ldots,E(t_a))\]
The head network ``decodes'' the embedding of the term considered in 
$\mathbb{R}^d$ into an element of the
output space $\mathbb{R}^n$.
Figure~\ref{fig:tnn} shows how the computation follows the tree structure of 
the input term.

In both experiments, the TNNs have neural network operators (including the head 
network) with one hidden layer and with a embedding dimension $d=12$.
we follow a training schedule over 200 epochs using a 
fixed learning rate of 0.02 and we double the batch size after every 50 epochs 
from 8 to 64.

\section{Arithmetical Expression Evaluation}
The aim of this task is to compute the value $x$ of a given
arithmetical expression. Since the output of the TNN is a fixed vector in 
$\mathbb{R}^n$, we restrict the objective to predicting the four binary 
digits of $x$ modulo 16. We say that a prediction is accurate if the four 
predicted real numbers rounded to the nearest integer corresponds
to the four binary digits.

The bottom-up architecture of the TNN is ideally suited for this task as it 
is a natural way to evaluate an expression. And since the knowledge of 
the structure of the formula is hard-coded in the tree structure, we expect the 
TNN to generalize well. 
The experiments rely on a training set of 11990 arithmetical expressions and
a testing set of 10180 arithmetical expressions. These expressions are 
constructed using the four operators $0,s,+$ and $\times$. The deepest subterms 
of the expressions are made of unary numbers between $0$ and $10$ (e.g. 
$s^8(0) + s(s^3 (0) \times s^2(0))$). 
For further inspection, the datasets are available in this 
repository\footnote{\url{https://github.com/barakeel/arithmetic_datasets}}.

\begin{table}
\centering
\vspace{-5mm}
\begin{tabular}{lcccc}
\toprule
Predictors &\phantom{small} & Train & \phantom{small} & Test\\
\midrule
NearestNeighbor~\cite{DBLP:journals/tit/CoverH67} && 100.0 && 11.7\\
LibLinear~\cite{DBLP:journals/jmlr/FanCHWL08} && \phantom{0}84.4 && 18.3\\
XGBoost~\cite{xgboost}  && \phantom{0}99.5 && 16.8 \\
\textsf{NMT}~\cite{DBLP:conf/nips/VaswaniSPUJGKP17short} && 100.0 && 77.2\\
Tree\holfour && 97.7 && 90.1\\
\bottomrule
\end{tabular}
\caption{Percentage of accurate predictions on different 
test sets}\label{tab:computation}
\vspace{-5mm}
\end{table}

In Table~\ref{tab:computation}, we compare the accuracy of our TNN predictor 
(Tree\holfour) with feature-based predictors. These predictors are quite 
successful in the 
premise selection task in ITP Hammers~\cite{hammers4qed}. 
We experiment with these predictors using a standard set of 
syntactical features, which consists of all the subterms of the arithmetical 
expressions. This requires almost no engineering. The accuracy 
of these predictors on the test set is only slightly better than random 
(6.25\%). The obvious reason is that it is 
challenging to compute the value of an expression by merely comparing its 
subterms with features of terms in the training set. This highlights the need 
for some feature engineering using these predictors.
As a final comparison, we test the deep learning recurrent 
neural model \textsf{NMT} with parameters taken from those shown as best in the 
informal-to-formal task~\cite{DBLP:conf/mkm/WangKU18}. 
This is a sequence-to-sequence model with attention, typically used 
for machine translation. We use prefix notation for representing the terms as 
sequences for NMT and used one class per output modulo 16 instead of the 
4-bit encoding as NMT performed better with those. 
Despite the perfect training accuracy, the testing 
accuracy of NMT is below the TNN.

\section{Propositional Truth Estimation}
The aim of this task is to teach a TNN to estimate if a propositional
formula is true or not. 
For this propositional task,  we re-use the benchmark created by the authors of 
\cite{nn-entailment} which can be downloaded from this 
repository~\footnote{\url{https://github.com/deepmind/logical-entailment-dataset}}.
There, each problem is of the form $A \Vdash? B$. To re-use the implication 
operator, we instead solve the equivalent task of determining if $A \Rightarrow 
B$ is universally true or not.

Moreover, the propositional formulas contain boolean variables and a direct 
representation in our TNN would create one neural network operator for each 
named variables (up to 25 in the dataset). Our solution is to encode all 
variables using two operators $x$ and $\mathit{prime}$. First, we index the 
variables according to their order of appearance in the formula. Second, 
each variable  $x_i$ is replaced by the term $\mathit{prime}^i(x)$.
Thus, the input formulas are now represented by terms built from the 
set of operators $\lbrace x,\mathit{prime},\Rightarrow,\neg,\vee,\wedge 
\rbrace$.

Table~\ref{tab:entail} compares the results of our TNNs (Tree\holfour) on the 
truth estimation
task with the best neural network architectures for this task.
The first three architectures are the best extracted from the table of results 
in \cite{nn-entailment}. 
The first one is a tree neural network similar to ours, which also indexes the 
variables. A significant difference is that 
we use the $prime$ operator to encode variables while they instead rely on data 
augmentation by permuting the indices of variables. The second one replaces 
feedforward networks by LSTMs. The third architecture bases its decision on 
simultaneously using multiple embeddings for boolean variables. That is why 
this architecture is named PossibleWorld. In contrast, the TopDown 
architecture~\cite{DBLP:conf/iclr/Chvalovsky19} inverts the structure of the 
TNNs, and combines the embedding of boolean variables (that are now outputs) 
using recurrent networks. 
The results on the test set demonstrate that our 
implementation of TNN is at least as good as the one in \cite{nn-entailment} as 
it beats it on every test set.
Overall, the more carefully designed architectures for this task (PossibleWorld 
and TopDown) outperform it. 
One thing to note is that these architectures 
typically rely on a much larger embedding dimension (up to $d=1024$ for the 
TopDown architecture). 
Our TNN implementation rivals with the best 
architectures on the exam dataset, which consists of 100 examples 
extracted from textbooks.

\begin{table}
\centering
\setlength{\tabcolsep}{1em}
\begin{tabular}{lccccc}
\toprule
Architecture & easy & hard & big & mass. & exam \\
\midrule
Tree &  72.2 & 69.7 & 67.9 & 56.6 & 85.0 \\
TreeLSTM &  77.8 & 74.2 & 74.2 & 59.3 & 75.0 \\
PossibleWorld & 98.6 & 96.7 & 93.9 & 
73.4 & 96.0\\
TopDown & 95.9 & 83.2 & 81.6 & 83.6 & 96.0\\
Tree\holfour & 86.5 & 77.8 & 79.2 & 61.2 & 98.0\\
\bottomrule
\end{tabular}
\caption{Percentage of accurate predictions}\label{tab:entail}
\vspace{-5mm}
\end{table}

\section{Usage}
Our deep learning modules allow HOL4 users to train a TNN on a chosen
supervised learning task with little development overhead. The
function \texttt{train\_tnn} from the module \texttt{mlTreeNeuralNetwork} is
available for such purpose. Its three arguments are a schedule, an initial TNN, 
and a couple consisting of training examples and testing examples. 

\paragraph{Examples}
Given the objective functions $o_1,\ldots,o_n$, an example for a term $t$ is:
  \[\lbrack(h_1(t),l_1),\ (h_2(t),l_2),\ldots,(h_n(t),\ l_n)\rbrack\]
where $l_i$ is the list of real numbers between 0 and 1 returned by $o_i(t)$ 
and $h_i$ is the head operator with objective $o_i$.
The term $t$ is expected to be lambda-free with each operator appearing with a 
unique arity. Each task in our experiments is defined by a single objective on 
a set of training examples.

\paragraph{Initial TNN}
To create an initial TNN, the user first needs to gather all operators 
appearing in the examples. Then, given an embedding dimension $d$, for each 
operator $f$ with arity $a$ the list of dimensions of $N_f$ is to be defined as:
  \[\lbrack a \times d, u_1,\ldots,u_k, d \rbrack\]
The natural numbers $u_1,\ldots,u_k$ are sizes of the intermediate layers
that can be freely chosen by the user. In the case of a head operator $h_i$, 
the input dimension is to be $d$ and the output dimension is to
be the length of the list $l_i$.
From the operators (including heads) and the associated dimensions, the user 
can randomly initialize the weights of the TNN by calling \texttt{random\_tnn}.

\paragraph{Schedule}
The schedule argument is a list of records containing hyperparameters for the 
training such as the number of threads, the number of epochs, the learning 
rate and the size of the batches. Here is a typical training schedule:
\begin{alltt}\small
[\{batch\_size = 16, learning\_rate = 0.02, ncore = 4, nepoch = 50, \ldots \},
 \{batch\_size = 32, learning\_rate = 0.02, ncore = 4, nepoch = 100, \ldots \}]
\end{alltt}
In this schedule, training is performed with a batch size of 16 for 50 epochs 
which is then increased to 32 for the next 100 epochs.

\section{Conclusion}
In this paper, we presented an implementation of tree neural networks(TNNs) in 
HOL4 that can be used to learn a function on HOL4 formulas from examples.
Compared to the other machine learning predictors, it excels on the 
arithmetical evaluation task as the TNN architecture reflects perfectly the 
implied bottom-up computation. 
It also exhibits excellent performance on propositional 
formulas. It yields a better accuracy than an existing implementation of TNNs 
but comes short of more involved architectures tailored for this particular 
task. As a way forward, we would like to see if the observed TNNs pattern 
recognition abilities (understanding) transfer to other tasks such as 
premise selection or high-order unification, which could have a more direct 
benefit for proof automation.

\bibliographystyle{splncs04}
\bibliography{biblio}
\end{document}